\documentclass[
]{ceurart}

\usepackage{listings}
\usepackage[T1]{fontenc}
\usepackage{graphicx}
\usepackage{subcaption}

\usepackage[T1]{fontenc}
\usepackage{adjustbox}
\usepackage{graphicx}
\usepackage{amsmath}
\usepackage{booktabs}
\usepackage{subcaption}
\usepackage{multirow}
\usepackage{ccicons}

\begin{document}
\conference{The 2nd World Conference on eXplainable Artificial Intelligence}

\title{Investigating Neuron Ablation in Attention Heads: The Case for Peak Activation Centering}

\copyrightyear{2024}
\copyrightclause{Copyright for this paper by its authors.
  Use permitted under Creative Commons License Attribution 4.0
  International (CC BY 4.0).}

\author[1]{Nicholas Pochinkov}[email=work@nicky.pro]
\address[1]{Independent. Dublin, Ireland}
\author[2]{Ben Pasero}
\author[2]{Skylar Shibayama}
\address[2]{Independent. Seattle, WA, USA}

\begin{abstract}
The use of transformer-based models is growing rapidly throughout society. With this growth, it is important to understand how they work, and in particular, how the attention mechanisms represent concepts. Though there are many interpretability
methods, many look at models through their neuronal activations, which
are poorly understood. We describe different lenses though which to view
neuron activations, and investigate the effectiveness in language models
and vision transformers though various methods of neural ablation: zero
ablation, mean ablation, activation resampling, and a novel approach
we term `peak ablation'. Through experimental analysis, we find that in
different regimes and models, each method can offer the lowest degradation of model performance compared to other methods, with resampling
usually causing the most significant performance deterioration. 
We make our code available at 
\hyperlink{https://github.com/nickypro/investigating-ablation}{https://github.com/nickypro/investigating-ablation}
\end{abstract}

\begin{keywords}
    AI \sep LLMs \sep Transformers \sep Interpretability \sep Attention \sep Pruning.
\end{keywords}

\maketitle

\section{Introduction}

Understanding how language models make decisions is important to ensure that their use can be trusted. Mechanistic interpretability offers one lens through which to understand how transformer architecture models \cite{attention-is-all-you-need} perform the computations required to get an output. An oft-used tool in mechanistic interpretability is to attribute individual network parts to specific capabilities by ablating those parts and observing capability degradation.

However, choosing how to ablate neurons in language models is still an unsolved problem. The traditional closed-form methods are zero ablation and mean ablation \cite{ablation-in-mnist,causal-scrubbing}, as well as an additional, more randomised method of activation resampling in the case of causal scrubbing \cite{causal-scrubbing,activation-patching-practices}, but little empirical analysis has been done to optimise these methods \cite{activation-patching-practices}.

Understanding exactly how neuron activations deviate, and what baseline they deviate from, is a broadly applicable question that is underexplored, and has the potential to improve techniques for model pruning and analysis into model sparsity.

In this paper, we 1) describe a simplistic working model of neuron activations, 2) suggest an improved, closed-form method of neuron ablation using modal activation, called `peak ablation', and 3) run experimental analysis on various ablation methods to compare the degree to which they harm model performance.

\section{Related Work}

\textbf{Mechanistic interpretability} is a field of research focusing on understanding how neural network models achieve their outputs. 
\cite{DBLP:conf/emnlp/GevaSBL21,causal-scrubbing,acdc,logit-lens,tunedlens,circuits-induction-heads}. 
A common method used in mechanistic interpretability, is `ablate and measure' \cite{causal-scrubbing}.
We investigate more precisely how different ablation methods affect performance, and propose `peak ablation' as another possible method.

Most relevantly, recent research \cite{activation-patching-practices} investigates hyperparameter selection to optimise activation patching
for causal scrubbing.
Our research differs; instead of interpolating activations between similar inputs, we set neurons' values for all inputs, and do not limit only to resampling.



\textbf{Pruning}: Model pruning \cite{neural-network-pruning-2020} is a common practice wherein reduced neural network parameter counts lessen memory and performance costs. In particular, structured pruning of large features 
\cite{structured-pruning} 
 is interested in the removal on the scale of neurons and attention heads, and can often achieve a large reduction in parameter count
\cite{pruning-sparse-gpt}
. Our work seeks to question the assumption of using masks that set neuron values to zero.

\textbf{Modularity:} Research into activation sparsity \cite{deja-vu-contextual-sparsity}, modularity \cite{emergent-modularity}, mixture of experts \cite{moefication,emergent-modularity,modular-deep-learning}, and unlearning by pruning \cite{pochinkov2023dissecting, selective-synaptic-dampening} all investigate how different subsets of activations are responsible for different tasks. These implicitly set activations to zero.


\section{Method}

\subsection{Pre-Trained Models and Datasets}
%
We work with two causal text models,
 Mistral 7B \cite{mistral} and
 Meta's OPT 1.3B \cite{meta-opt},
 a masked text model,
RoBERTa Large \cite{roberta}, and a vision transformer,
ViT Base Patch16 224 \cite{vision-transformer-google-2021}.


To get a general sense of performance, the above models were evaluated by looking at top1 prediction accuracy\footnote{top1 token prediction accuracy for language models, top1 image classification accuracy for image models}, as well as cross-entropy loss on various datasets.
For text models, we assess on EleutherAI's `The Pile' \cite{pile}. For image models, we assess on Imagenet-1k \cite{imagenet}, an image dataset with 1000 different classes. We evaluate on deterministic subsets of 100,000 text tokens and 1000 images respectively

\subsection{Neurons}\label{sec:neurons}

The objects of study are attention pre-out neurons, sometimes called `z'-hook activations.
We define attention pre-out neuron activations 
$y_i=f(x_i)=\text{preout}(x_i)$ as
$y_i = \sum_j A_{i,j} W_V x_j $, where 
$A_{i,j} = \text{softmax}( (W_K x_i) \cdot (W_Q x_j) )$, where $W_Q, W_K, W_V$ are the attention query, key, and value matrices respectively. 
We focus on attention neurons rather than MLP neurons, as these do not have an activation function that privileges positive activations, making analysis more difficult.
To ablate a neuron, we replace $y_i = f(x_i)$ with some constant.

\begin{figure}[h]
    \centering
    \includegraphics[width=0.2\linewidth]{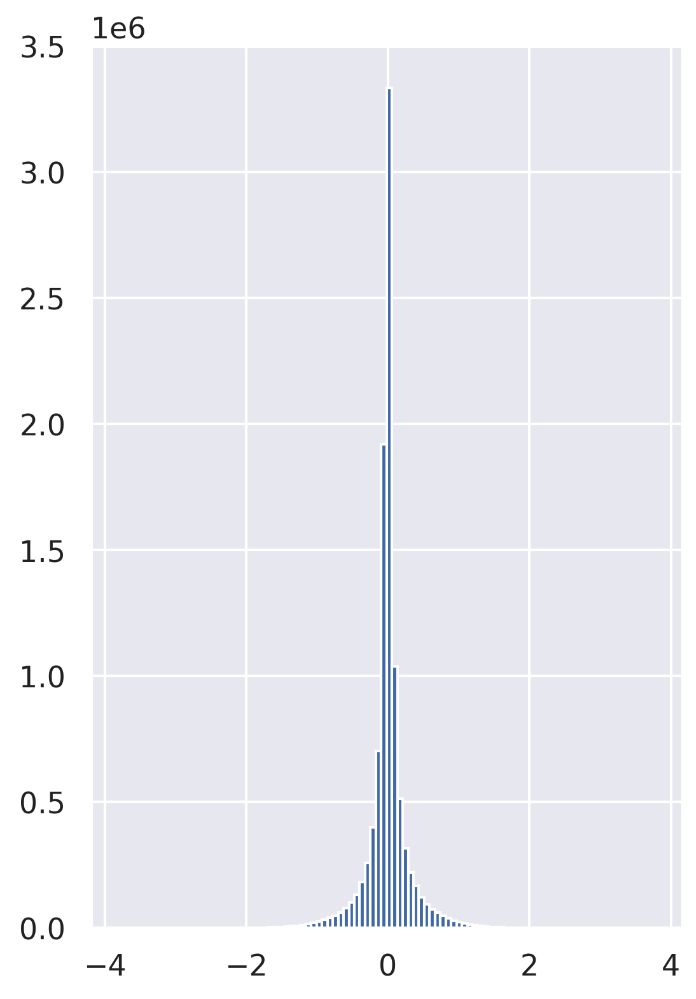}
    \hspace{0.05\linewidth}
    \includegraphics[width=0.2\linewidth]{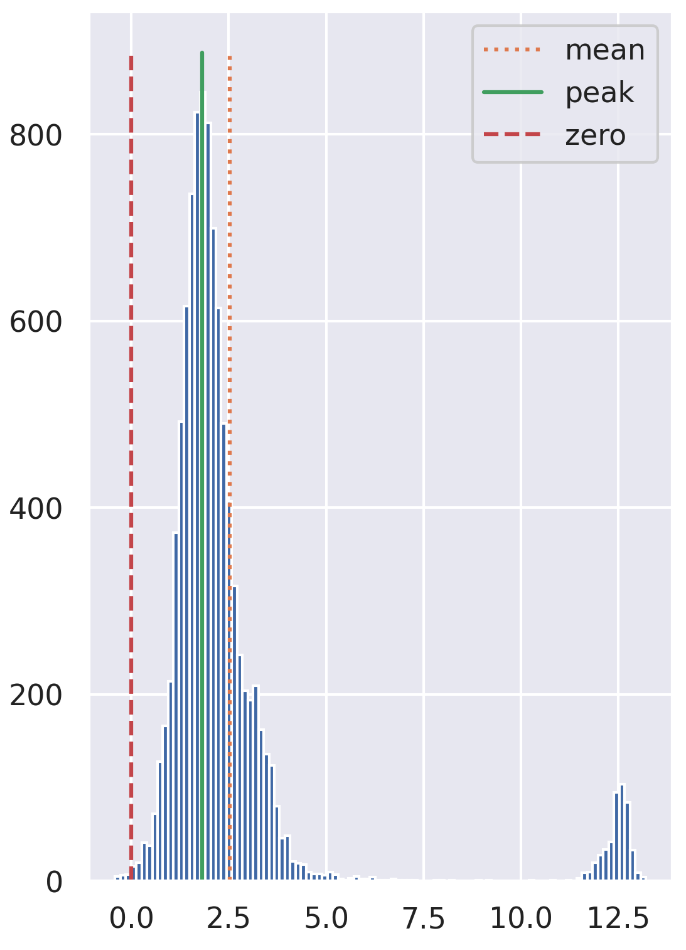}
    \includegraphics[width=0.2\linewidth]{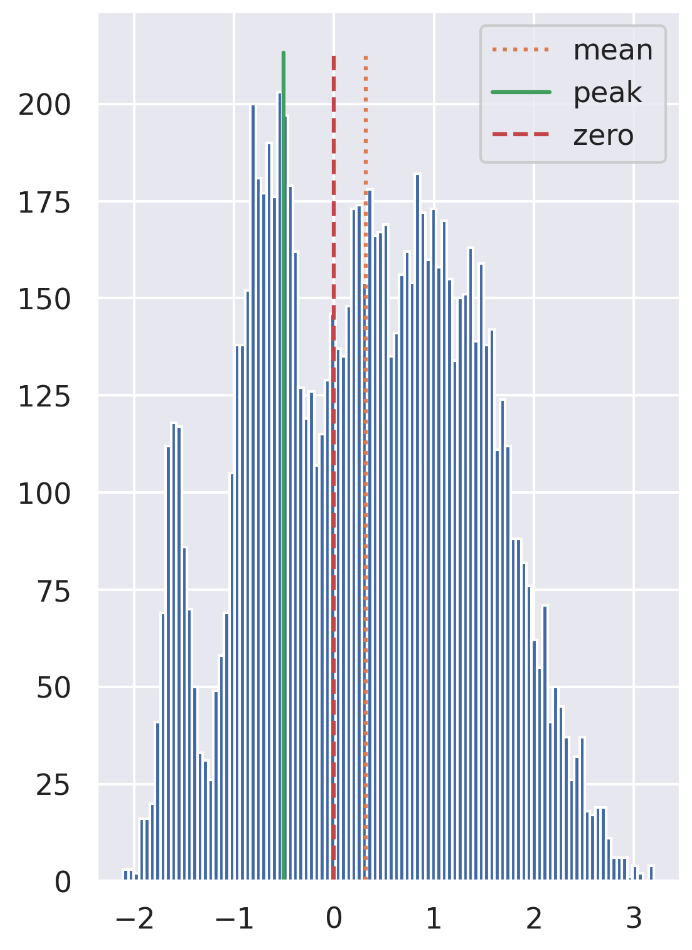}
    \caption{Un-normalised probability density functions (histograms) of attention neuron activations in RoBERTa. We see in (left) an average of distributions of all neurons in a layer, (centre) a bi-modal neuron with both peaks not at zero, and (right) another example of a neuron with an atypical distribution. X-axis shows neuron value, and Y-axis shows probability of a neuron taking that value.}
    \label{fig:results-neuron-plots}
    \vspace{-12px}
\end{figure}

In Figure \ref{fig:results-neuron-plots}, we showcase some plots of neuron probability distributions. We see an example of many attention pre-out neuron activation distributions within the same layer. We note that most neurons follow a roughly Gaussian or double-exponential distribution about zero, but note that there is a minority of neurons that are not distributed at zero. As most neurons are zero-centred and symmetric, it makes sense that zero and mean ablation work quite well.

\subsection{A Working Model of Neuron Activation}\label{sec:neuron-model}
Our hypothesis, based on activation profiles such as those seen in
Figure \ref{fig:results-neuron-plots}
is that neurons have a `baseline' or `default-mode' activation (typically at zero), when the input contains no relevant features, which is then deviated from as neurons fire in proportion to various features they are tuned to pick up.
In residual stream models \cite{resnet-original-paper}, information is limited to the width of the residual stream \cite{residual-deep-limits}, and as the residual stream typically grows exponentially in size \cite{residual-grows}, noise can become amplified.
This is supported by the common redundancy of many circuits \cite{circuits-ioi-interpretability-redundancy}, even in transformer models trained without the use of dropout \cite{hydra-effect}.

In particular, we expect that ablating neurons should have two contributors to reduced performance. These are 1) removing the relevant contextual computed information that the neuron is providing, and 2) taking the model activation out of distribution, by adding `noise'. Ablating neurons to a constant value should cause some constant increase in loss for term 1, and different constant should contribute to different values of term 2. As we increase the distance from the `default-mode' value, the neuron would further degrade the performance by taking the residual stream further out of distribution, thus in some sense, `adding noise'. 

\subsection{Ablation Methods}
We choose four main methods of ablating neurons, see Table \ref{tab:ablation-methods} for a summary. These are:

\begin{table*}[h]
    \caption{Comparison between the neural ablation methods described.}
    \label{tab:ablation-methods}
    \centering
    \begin{tabular}{ll} 
    \toprule
    Method & Set the neuron activation... \\
    \midrule
    Zero Ablation         & ...to zero \\
    Mean Ablation         & ...to the mean value within the dataset D \\
    Activation Resampling & ...to some values from some different input \\ 
    Naive Peak Ablation   & ...to the modal `peak' activation value within the dataset D \\
    \bottomrule
    \end{tabular}
\end{table*}

\textbf{Zero ablation}: The most common form of ablation, which involves replacing a neuronal activation of any with a zeroed out activation. That is, setting $\forall x_j:  f_i(x_j) = 0.0$

\textbf{Mean Ablation}: A still relatively-common method of ablation, which involves first collecting activations of various neurons on a distribution of inputs, and averaging the activations to find a mean activation. That is, for some dataset $D$, for $x_j \in D$, let $f_i(x_j) = \frac{1}{|D|} \sum_j f_i(x_j) $

\textbf{Activation Resampling}: Inspired by \cite{causal-scrubbing,activation-patching-practices}, we also try general neuron resampling, by setting activations to those found by giving a randomised input. 
\footnote{This differs slightly to the original description, as in other research, they use a specific task, like circuit analysis \cite{causal-scrubbing} for the activation resampling, where the specific prompt template already exists.}
For text model, we take activations by a) sampling random generated characters, b) sampling random tokens, and c) using OPT to generate a random text. For ViT, we use randomly generated pixel values.

\textbf{Naive Peak Ablation}: Observing that neuronal activations frequently exhibit a prominent peak, we propose an ablation method targeting their modal activation. For bin size \(\epsilon\), the neuron $i$ activations \(f_i(x_j)\) for each \(x_j \in D\) are sorted into bins \(N_i[k]\) such that \(y_k \leq f_i(x_j) < y_k + \epsilon\). The bin \(N_i[k_{max}]\) with the highest occurrence is selected, and \(f_i(x_j)\) is set to \(y_{k_{max}} + \frac{\epsilon}{2}\).

\subsection{Ablation Experiments}
Under the working model described in Section \ref{sec:neuron-model}, we expect that ablating neurons to different values should have different impacts to performance, with there being a value which leads to some minimal drop in performance due to minimal noise being added to the residual stream. 

We randomly select attention neurons in increments of 10\% and ablate them until the model is fully pruned, and at each step, assess performance by evaluating the Top1 accuracy and Cross-Entropy Loss in the chosen dataset with each ablation method, described in
Table \ref{tab:ablation-methods}.
The neurons are selected deterministically across three separate seeds, summarised in Table \ref{tab:comp}

\section{Results}

\subsection{Causal Text Models}

\begin{figure}[h]
\centering
\begin{subfigure}[b]{0.245\textwidth}
    \centering
    \includegraphics[width=\linewidth]{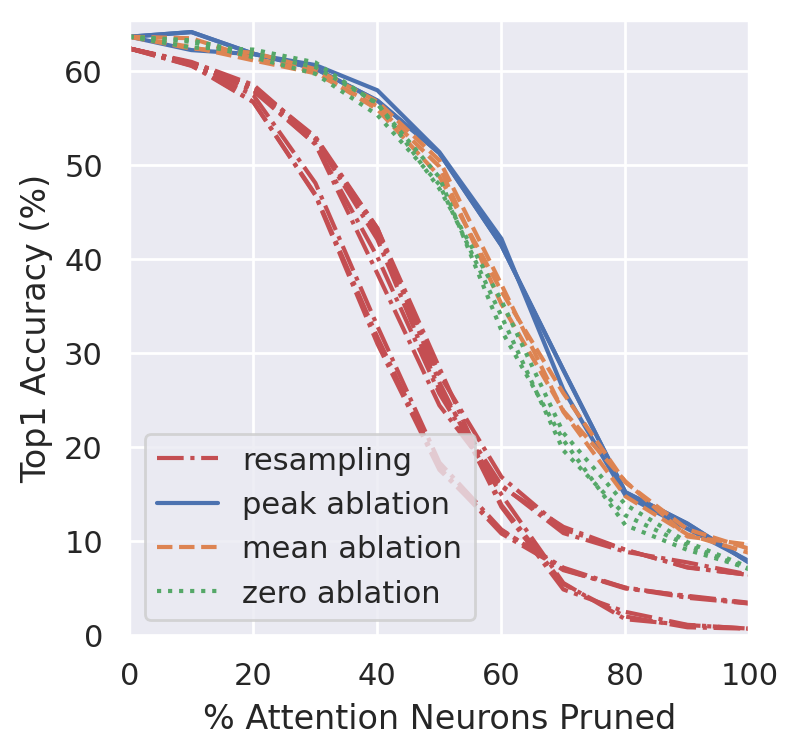}
    \caption{Mistral 7B Top1}
    \label{fig:mistral-top1}
\end{subfigure}
\hfill
\begin{subfigure}[b]{0.245\textwidth}
    \centering
    \includegraphics[width=\linewidth]{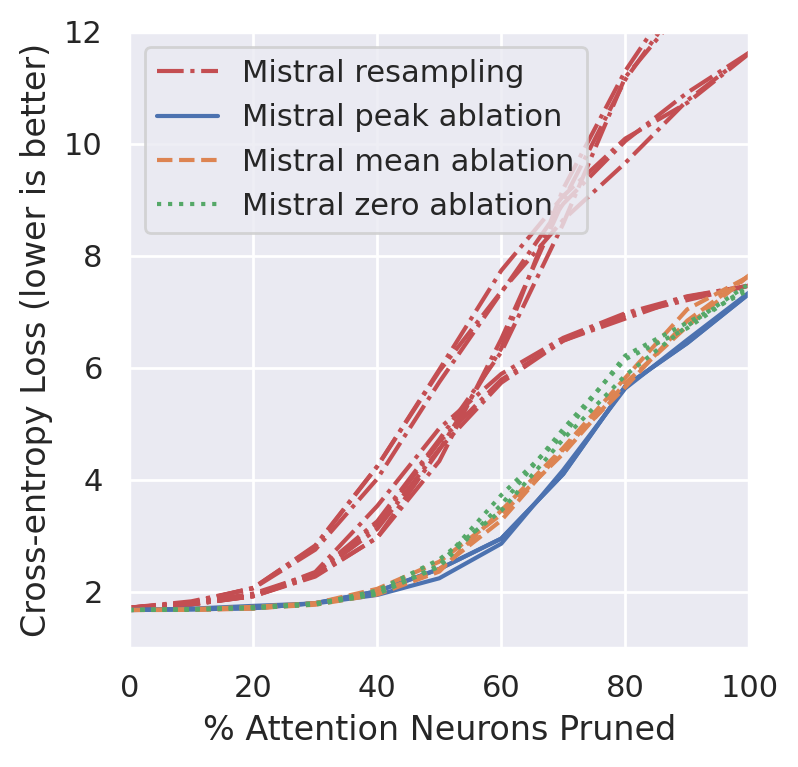}
    \caption{Mistral 7B CE Loss}
    \label{fig:mistral-ce}
\end{subfigure}
\hfill
\begin{subfigure}[b]{0.245\textwidth}
    \centering
    \includegraphics[width=\linewidth]{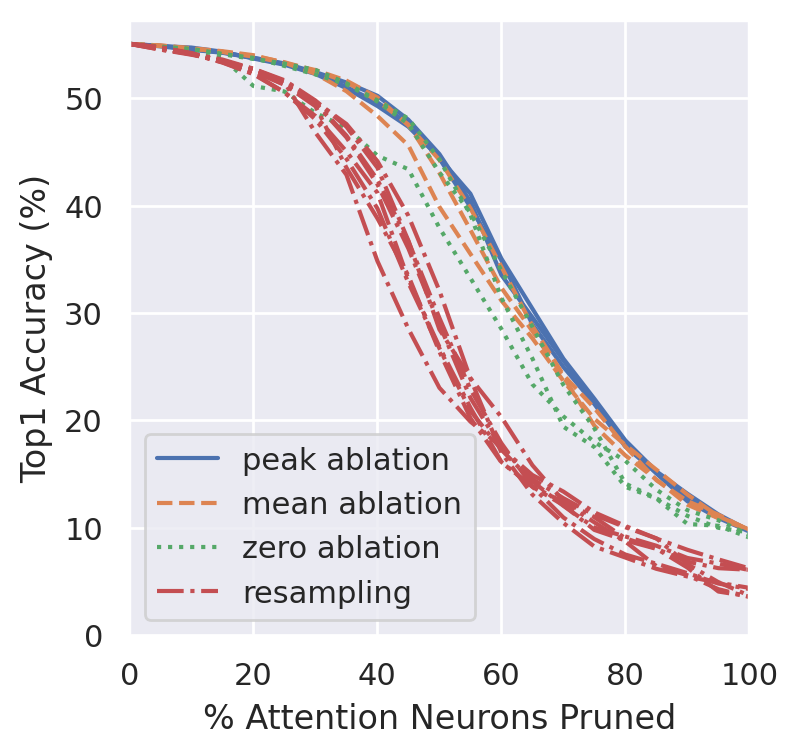}
    \caption{OPT 1.3B Top1}
    \label{fig:opt-top1}
\end{subfigure}
\hfill
\begin{subfigure}[b]{0.245\textwidth}
    \centering
    \includegraphics[width=\linewidth]{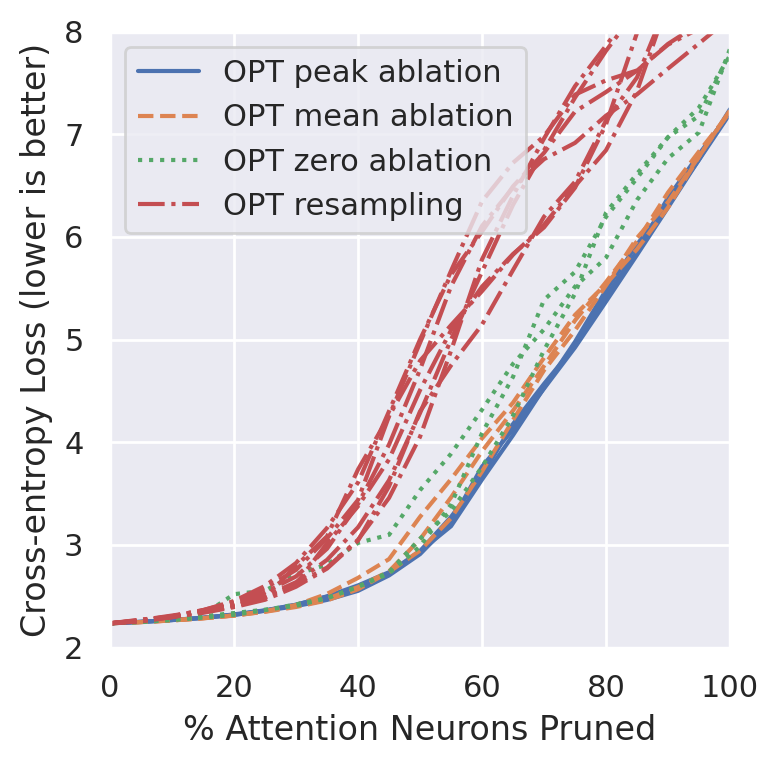}
    \caption{Opt 1.3B CE Loss}
    \label{fig:opt-ce}
\end{subfigure}
\vspace{-12px}
\caption{Change in Top1 next-token prediction accuracy (Top1) and cross-entropy loss (CE Loss) at different fractions of model pruned with different methods of ablation for Mistral 7B and OPT 1.3B}
\label{fig:random-compare}
\vspace{-10px}
\end{figure}

In Figure \ref{fig:random-compare}, we see the results for random pruning of OPT 1.3b and Mistral 7b with the different methods of ablation. We can see that Peak ablation has the most consistent pattern, causing the lowest amount of degradation, with mean ablation and zero ablation coming a close second and third, and Random resampling causes by far the most degradation.
Of the three resampling methods, choosing random tokens causes the lowest degradation.

\subsection{Other Transformers}

\begin{figure}[h]
\centering
\begin{subfigure}[b]{0.245\textwidth}
    \centering
    \includegraphics[width=\linewidth]{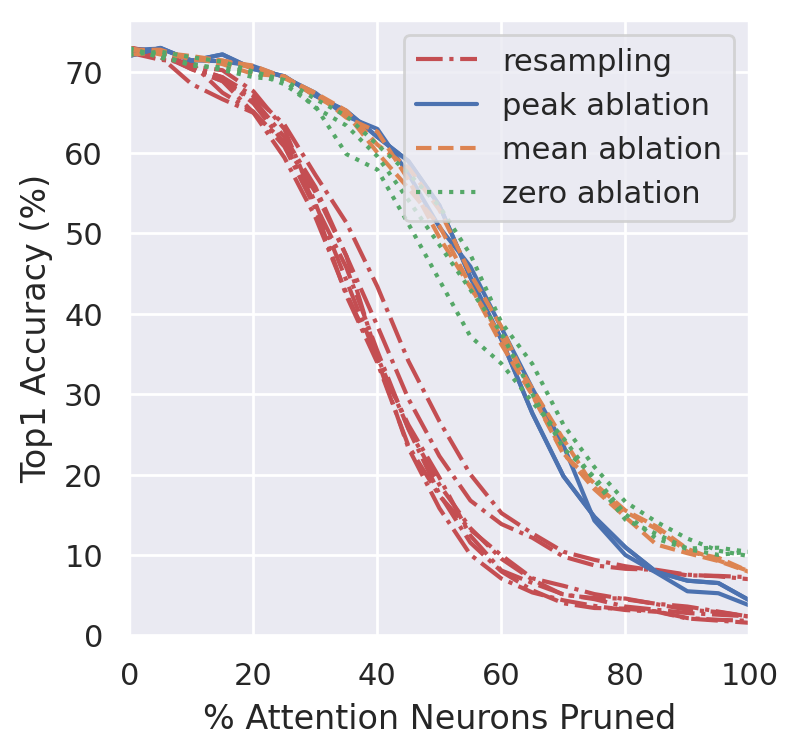}
    \caption{RoBERTa Top1}
    \label{fig:roberta-top1}
\end{subfigure}
\hfill
\begin{subfigure}[b]{0.245\textwidth}
    \centering
    \includegraphics[width=\linewidth]{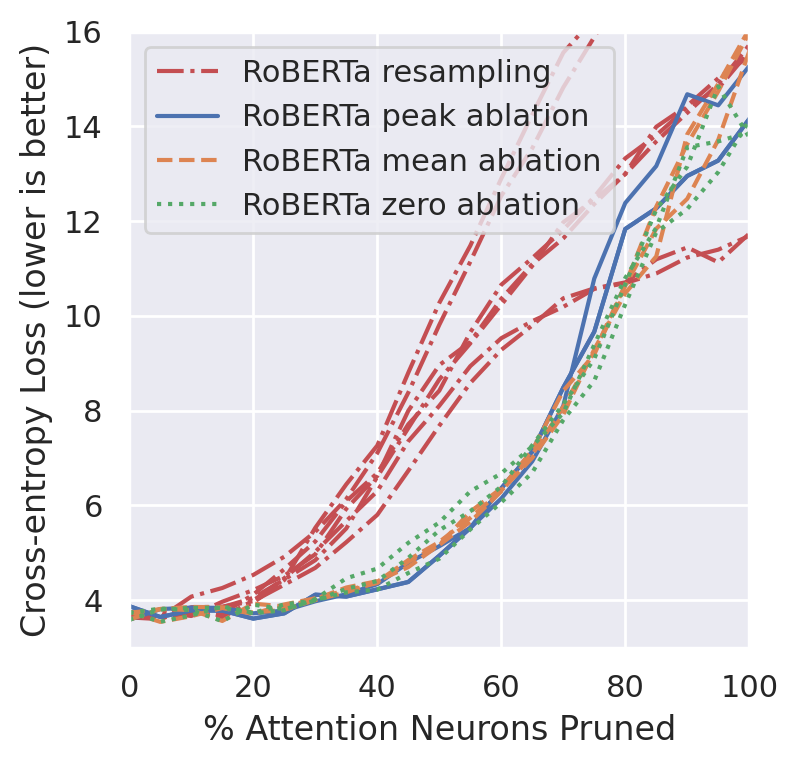}
    \caption{RoBERTa CE Loss}
    \label{fig:vit-top1}
\end{subfigure}
\hfill
\begin{subfigure}[b]{0.245\textwidth}
    \centering
    \includegraphics[width=\linewidth]{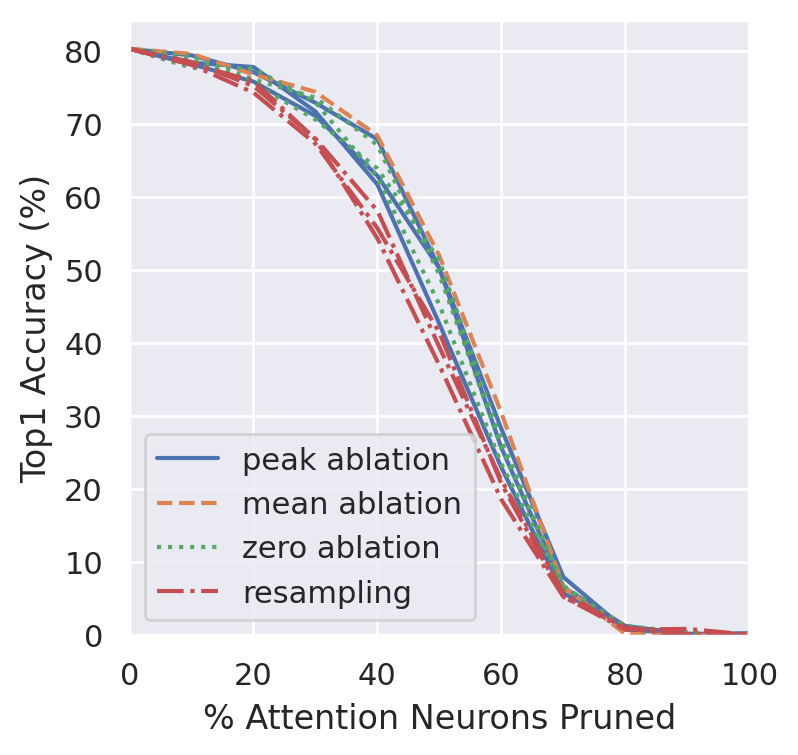}
    \caption{ViT Top1}
    \label{fig:roberta-ce}
\end{subfigure}
\hfill
\begin{subfigure}[b]{0.245\textwidth}
    \centering
    \includegraphics[width=\linewidth]{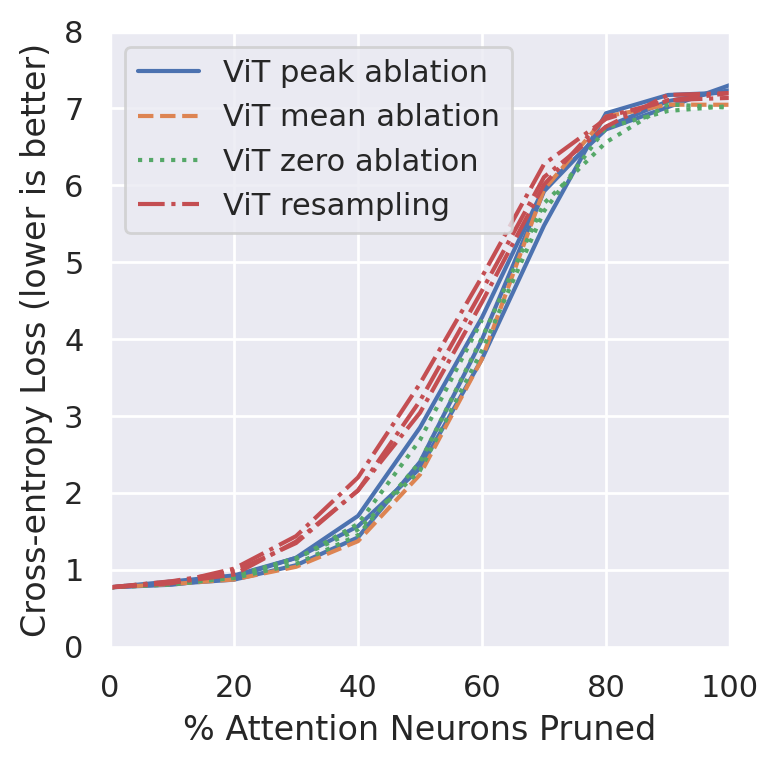}
    \caption{ViT CE Loss}
    \label{fig:vit-ce}
\end{subfigure}
\vspace{-12px}
\caption{Change in Top1 next-token prediction accuracy (Top1) and cross-entropy loss (CE Loss) at different fractions of model pruned with different methods of ablation for ViT 7B and RoBERTa}
\label{fig:random-compare-other}
\vspace{-10px}
\end{figure}

In Figure \ref{fig:random-compare-other}, we see that for ViT, zero, mean, and peak ablation have statistically insignificant differences in performance, while resampling causes some small additional degradation. We can see that almost all of the performance loss is based on the specific neurons being selected rather than the ablation method being chosen, even between zero, mean, and peak ablation.

In RoBERTa, we see that in the first 75\% of pruning, the three main methods of peak, mean and zero ablation are very close, with Peak edging slightly better performance. Beyond 75\%, the three methods become more noisy; resampling of IDs ends up having the best performance overall in both Top1 and Cross-Entropy Loss at the task of token unmasking and de-randomization.

\subsection{Overall Comparison}
 In Table \ref{tab:comp}, we see that in different models and in different regimes, the different methods have different merits in reducing performance, with Peak ablation working overall best in the most cases.
 Surprisingly, although Random resampling seems to add a lot of noise to the activations, random token ID resampling can sometimes work well, such as in RoBERTa. 

\newcommand{\mc}[1]{\multicolumn{2}{c|}{#1}}
\newcommand{\mf}[1]{\multicolumn{5}{c}{#1}}
\newcommand{\mr}[1]{\multirow{6}{*}{#1}}

\begin{table*}
\caption{Performance impact of neural ablation methods on the attention neurons of OPT, Mistral, ViT and RoBERTa. Ablation methods are Peak, Mean and Zero ablation, as well as Resampling (RS) with random characters (RS1), token IDs (RS2) and generated text (RS3) for text models, and random pixels (RS1) for ViT. Models are pruned by randomly selecting $50\%$ and $90\%$ of neurons}
\vspace{4px}
\label{tab:comp}
\centering
\begin{adjustbox}{center}
\begin{tabular}{cc | rrrr }
\toprule
\mc{Top1 Accuracy} & OPT & Mistral & ViT & RoBERTa \\
\midrule
\mc{Baseline} & $55.05 \pm 0.00$ & $60.05 \pm 0.00$ & $80.32 \pm 0.00$ & $73.04 \pm 0.00$ \\
\midrule
& Peak & $\mathbf{44.54\pm0.19}$ & $\mathbf{51.30\pm0.05}$ & $\mathbf{47.72\pm3.58}$ & $\mathbf{52.55\pm1.27}$ \\
& Mean & $42.39\pm1.84$ & $49.71\pm0.70$ & $\mathbf{48.65\pm5.38}$ & $51.10\pm1.41$ \\
50 \%
& Zero & $41.77\pm2.76$ & $48.03\pm0.50$ & $\mathbf{48.65\pm2.54}$ & $48.69\pm3.75$ \\
Pruning
& RS1 & $25.97\pm2.24$ & $27.01\pm1.12$ & $39.29\pm1.84$ & $18.57\pm0.93$ \\
& RS2 & $27.51\pm1.02$ & $26.10\pm1.34$ & - & $24.83\pm1.84$ \\
& RS3 & $29.90\pm1.65$ & $17.93\pm0.33$ & - & $16.10\pm1.12$ \\
\midrule
& Peak & $\mathbf{12.81\pm0.29}$ & $\mathbf{11.70\pm0.26}$ & $0.20\pm0.00$ & $6.37\pm0.62$ \\
& Mean & $12.59\pm0.39$ & $10.84\pm0.32$ & $0.17\pm0.17$ & $10.46\pm0.22$ \\
90\%
& Zero & $11.05\pm0.53$ & $9.53\pm0.34$ & $\mathbf{0.50\pm0.37}$ & $\mathbf{11.20\pm0.60}$ \\
Pruning
& RS1 & $6.18\pm0.43$ & $1.03\pm0.12$ & $\mathbf{0.37\pm0.39}$ & $3.19\pm0.33$ \\
& RS2 & $7.35\pm0.46$ & $7.41\pm0.25$ & - & $7.55\pm0.06$ \\
& RS3 & $5.55\pm0.17$ & $4.11\pm0.09$ & - & $2.26\pm0.18$ \\
\midrule
\mc{Cross-Entopy Loss} & OPT & Mistral & ViT & RoBERTa \\
\midrule
\mc{ Baseline } & $2.24 \pm 0.00$ & $1.89 \pm 0.00$ & $0.77 \pm 0.00$ & $3.75 \pm 0.00$ \\
\midrule
& Peak & $\mathbf{2.93\pm0.01}$ & $\mathbf{2.35\pm0.08}$ & $\mathbf{2.52\pm0.23}$ & $\mathbf{5.00\pm0.09}$ \\
& Mean & $3.09\pm0.14$ & $2.43\pm0.07$ & $\mathbf{2.47\pm0.33}$ & $5.19\pm0.03$ \\
50\%
& Zero & $3.19\pm0.24$ & $2.49\pm0.06$ & $\mathbf{2.46\pm0.17}$ & $5.33\pm0.33$ \\
Pruning
& RS1 & $4.53\pm0.20$ & $4.52\pm0.16$ & $3.22\pm0.16$ & $8.68\pm0.22$ \\
& RS2 & $4.89\pm0.13$ & $4.71\pm0.15$ & - & $7.85\pm0.18$ \\
& RS3 & $4.26\pm0.16$ & $5.88\pm0.10$ & - & $10.09\pm0.20$ \\
\midrule
& Peak & $\mathbf{6.33\pm0.04}$ & $\mathbf{6.45\pm0.03}$ & $7.10\pm0.06$ & $13.53\pm0.81$ \\
& Mean & $\mathbf{6.35\pm0.06}$ & $6.87\pm0.12$ & $7.07\pm0.04$ & $13.31\pm0.60$ \\
90\%
& Zero & $6.90\pm0.10$ & $6.75\pm0.04$ & $\mathbf{6.99\pm0.04}$ & $12.99\pm0.54$ \\
Pruning
& RS1 & $8.40\pm0.15$ & $12.71\pm0.18$ & $7.13\pm0.03$ & $14.37\pm0.06$ \\
& RS2 & $7.80\pm0.11$ & $7.25\pm0.03$ & - & $\mathbf{11.32\pm0.09}$ \\
& RS3 & $8.67\pm0.09$ & $10.80\pm0.08$ & - & $20.85\pm0.19$ \\
\bottomrule
\end{tabular}
\end{adjustbox}
\vspace{-10px}
\end{table*}

%

\section{Discussion}

The analysis presented seems to suggest that when evaluating and understanding neurons in the attention layers of language models, the ideal centring method seems to depend significantly on the model. In decoder models, a good method is to find the largest peak, with a close second being zero ablation. This similarity is expected, as most neurons are centred at zero. This has downstream effects on improving the way we can look at one of the most crucial aspects of how neural networks work - their activations.

We have seen that neurons can have activations that are: non-Gaussian, non-symmetric, multi-modal, non-zero-centred. We hypothesise that taking into consideration this fact has the potential to make interpretability analysis into more fruitful, and centring activations by their peak seems a potential natural method.

Future work could: 1) investigate other potential better methods for neuron recentring, 2) more thoroughly investigate the differences between `well-behaved' symmetric zero-centred distributions, and those that deviate from this norm, 3) find more efficient ways of computing the peak activations for larger models.




\section{Acknowledgements} 
This research was funded by the Long-Term Future Fund, and done in affiliation with AI Safety Camp.

\bibliography{references}

\begin{thebibliography}{29}
\expandafter\ifx\csname natexlab\endcsname\relax\def\natexlab#1{#1}\fi
\providecommand{\url}[1]{\texttt{#1}}
\providecommand{\href}[2]{#2}
\providecommand{\path}[1]{#1}
\providecommand{\DOIprefix}{doi:}
\providecommand{\ArXivprefix}{arXiv:}
\providecommand{\URLprefix}{URL: }
\providecommand{\Pubmedprefix}{pmid:}
\providecommand{\doi}[1]{\href{http://dx.doi.org/#1}{\path{#1}}}
\providecommand{\Pubmed}[1]{\href{pmid:#1}{\path{#1}}}
\providecommand{\bibinfo}[2]{#2}
\ifx\xfnm\relax \def\xfnm[#1]{\unskip,\space#1}\fi
\bibitem[{Vaswani et~al.(2017)Vaswani, Shazeer, Parmar, Uszkoreit, Jones,
  Gomez, Kaiser, and Polosukhin}]{attention-is-all-you-need}
\bibinfo{author}{A.~Vaswani}, \bibinfo{author}{N.~Shazeer},
  \bibinfo{author}{N.~Parmar}, \bibinfo{author}{J.~Uszkoreit},
  \bibinfo{author}{L.~Jones}, \bibinfo{author}{A.~N. Gomez},
  \bibinfo{author}{{\L}.~Kaiser}, \bibinfo{author}{I.~Polosukhin},
\newblock \bibinfo{title}{Attention is all you need},
\newblock \bibinfo{journal}{Advances in neural information processing}
  \bibinfo{volume}{30} (\bibinfo{year}{2017}).
\bibitem[{Meyes et~al.(2019)Meyes, Lu, de~Puiseau, and
  Meisen}]{ablation-in-mnist}
\bibinfo{author}{R.~Meyes}, \bibinfo{author}{M.~Lu}, \bibinfo{author}{C.~W.
  de~Puiseau}, \bibinfo{author}{T.~Meisen},
\newblock \bibinfo{title}{Ablation studies in artificial neural networks},
\newblock \bibinfo{journal}{arXiv preprint arXiv:1901.08644}
  (\bibinfo{year}{2019}).
\bibitem[{Chan et~al.(2022)Chan, Garriga-Alonso, Goldowsky-Dill, Greenblatt,
  Nitishinskaya, Radhakrishnan, Shlegeris, and Thomas}]{causal-scrubbing}
\bibinfo{author}{L.~Chan}, \bibinfo{author}{A.~Garriga-Alonso},
  \bibinfo{author}{N.~Goldowsky-Dill}, \bibinfo{author}{R.~Greenblatt},
  \bibinfo{author}{J.~Nitishinskaya}, \bibinfo{author}{A.~Radhakrishnan},
  \bibinfo{author}{B.~Shlegeris}, \bibinfo{author}{N.~Thomas},
  \bibinfo{title}{Causal scrubbing: a method for rigorously testing
  interpretability hypotheses}, \bibinfo{howpublished}{AI Alignment Forum},
  \bibinfo{year}{2022}. \URLprefix
  \url{https://www.alignmentforum.org/posts/JvZhhzycHu2Yd57RN}.
\bibitem[{Zhang and Nanda(2023)}]{activation-patching-practices}
\bibinfo{author}{F.~Zhang}, \bibinfo{author}{N.~Nanda},
\newblock \bibinfo{title}{Towards best practices of activation patching in
  language models: Metrics and methods},
\newblock \bibinfo{journal}{arXiv preprint arXiv:2309.16042}
  (\bibinfo{year}{2023}).
\bibitem[{Geva et~al.(2021)Geva, Schuster, Berant, and
  Levy}]{DBLP:conf/emnlp/GevaSBL21}
\bibinfo{author}{M.~Geva}, \bibinfo{author}{R.~Schuster},
  \bibinfo{author}{J.~Berant}, \bibinfo{author}{O.~Levy},
\newblock \bibinfo{title}{Transformer feed-forward layers are key-value
  memories},
\newblock in: \bibinfo{editor}{M.~Moens}, \bibinfo{editor}{X.~Huang},
  \bibinfo{editor}{L.~Specia}, \bibinfo{editor}{S.~W. Yih} (Eds.),
  \bibinfo{booktitle}{Proceedings of the 2021 Conference on Empirical Methods
  in Natural Language Processing, {EMNLP} 2021, Virtual Event / Punta Cana,
  Dominican Republic, 7-11 November, 2021}, \bibinfo{publisher}{Association for
  Computational Linguistics}, \bibinfo{year}{2021}, pp.
  \bibinfo{pages}{5484--5495}.
\bibitem[{Conmy et~al.(2023)Conmy, Mavor{-}Parker, Lynch, Heimersheim, and
  Garriga{-}Alonso}]{acdc}
\bibinfo{author}{A.~Conmy}, \bibinfo{author}{A.~N. Mavor{-}Parker},
  \bibinfo{author}{A.~Lynch}, \bibinfo{author}{S.~Heimersheim},
  \bibinfo{author}{A.~Garriga{-}Alonso},
\newblock \bibinfo{title}{Towards automated circuit discovery for mechanistic
  interpretability},
\newblock \bibinfo{journal}{CoRR} \bibinfo{volume}{abs/2304.14997}
  (\bibinfo{year}{2023}). \href{http://arxiv.org/abs/2304.14997}{{\tt
  arXiv:2304.14997}}.
\bibitem[{nostalgebraist(2020)}]{logit-lens}
\bibinfo{author}{nostalgebraist},
\newblock \bibinfo{title}{interpreting gpt: the logit lens},
\newblock \bibinfo{journal}{LessWrong}  (\bibinfo{year}{2020}). \URLprefix
  \url{https://www.lesswrong.com/posts/AcKRB8wDpdaN6v6ru/interpreting-gpt-the-logit-lens}.
\bibitem[{Belrose et~al.(2023)Belrose, Furman, Smith, Halawi, Ostrovsky,
  McKinney, Biderman, and Steinhardt}]{tunedlens}
\bibinfo{author}{N.~Belrose}, \bibinfo{author}{Z.~Furman},
  \bibinfo{author}{L.~Smith}, \bibinfo{author}{D.~Halawi},
  \bibinfo{author}{I.~Ostrovsky}, \bibinfo{author}{L.~McKinney},
  \bibinfo{author}{S.~Biderman}, \bibinfo{author}{J.~Steinhardt},
\newblock \bibinfo{title}{Eliciting latent predictions from transformers with
  the tuned lens},
\newblock \bibinfo{journal}{arXiv preprint arXiv:2303.08112}
  (\bibinfo{year}{2023}).
\bibitem[{Olsson et~al.(2022)Olsson, Elhage, Nanda, Joseph, DasSarma, Henighan,
  Mann, Askell, Bai, Chen et~al.}]{circuits-induction-heads}
\bibinfo{author}{C.~Olsson}, \bibinfo{author}{N.~Elhage},
  \bibinfo{author}{N.~Nanda}, \bibinfo{author}{N.~Joseph},
  \bibinfo{author}{N.~DasSarma}, \bibinfo{author}{T.~Henighan},
  \bibinfo{author}{B.~Mann}, \bibinfo{author}{A.~Askell},
  \bibinfo{author}{Y.~Bai}, \bibinfo{author}{A.~Chen}, et~al.,
\newblock \bibinfo{title}{In-context learning and induction heads},
\newblock \bibinfo{journal}{arXiv preprint arXiv:2209.11895}
  (\bibinfo{year}{2022}).
\bibitem[{Blalock et~al.(2020)Blalock, Ortiz, Frankle, and
  Guttag}]{neural-network-pruning-2020}
\bibinfo{author}{D.~W. Blalock}, \bibinfo{author}{J.~J.~G. Ortiz},
  \bibinfo{author}{J.~Frankle}, \bibinfo{author}{J.~V. Guttag},
\newblock \bibinfo{title}{What is the state of neural network pruning?},
\newblock in: \bibinfo{editor}{I.~S. Dhillon}, \bibinfo{editor}{D.~S.
  Papailiopoulos}, \bibinfo{editor}{V.~Sze} (Eds.),
  \bibinfo{booktitle}{Proceedings of Machine Learning and Systems 2020, MLSys
  2020, Austin, TX, USA, March, 2020}, \bibinfo{publisher}{mlsys.org},
  \bibinfo{year}{2020}.
\bibitem[{Wang et~al.(2020)Wang, Wohlwend, and Lei}]{structured-pruning}
\bibinfo{author}{Z.~Wang}, \bibinfo{author}{J.~Wohlwend},
  \bibinfo{author}{T.~Lei},
\newblock \bibinfo{title}{Structured pruning of large language models},
\newblock in: \bibinfo{editor}{B.~Webber}, \bibinfo{editor}{T.~Cohn},
  \bibinfo{editor}{Y.~He}, \bibinfo{editor}{Y.~Liu} (Eds.),
  \bibinfo{booktitle}{Proceedings of the 2020 Conference on Empirical Methods
  in Natural Language Processing, {EMNLP} 2020, Online, November 16-20, 2020},
  \bibinfo{publisher}{Association for Computational Linguistics},
  \bibinfo{year}{2020}, pp. \bibinfo{pages}{6151--6162}.
\bibitem[{Frantar and Alistarh(2023)}]{pruning-sparse-gpt}
\bibinfo{author}{E.~Frantar}, \bibinfo{author}{D.~Alistarh},
\newblock \bibinfo{title}{Sparsegpt: Massive language models can be accurately
  pruned in one-shot},
\newblock in: \bibinfo{booktitle}{International Conference on Machine
  Learning}, \bibinfo{organization}{PMLR}, \bibinfo{year}{2023}, pp.
  \bibinfo{pages}{10323--10337}.
\bibitem[{Liu et~al.(2023)Liu, Wang, Dao, Zhou, Yuan, Song, Shrivastava, Zhang,
  Tian, R{\'{e}}, and Chen}]{deja-vu-contextual-sparsity}
\bibinfo{author}{Z.~Liu}, \bibinfo{author}{J.~Wang}, \bibinfo{author}{T.~Dao},
  \bibinfo{author}{T.~Zhou}, \bibinfo{author}{B.~Yuan},
  \bibinfo{author}{Z.~Song}, \bibinfo{author}{A.~Shrivastava},
  \bibinfo{author}{C.~Zhang}, \bibinfo{author}{Y.~Tian},
  \bibinfo{author}{C.~R{\'{e}}}, \bibinfo{author}{B.~Chen},
\newblock \bibinfo{title}{Deja vu: Contextual sparsity for efficient llms at
  inference time},
\newblock in: \bibinfo{editor}{A.~Krause}, \bibinfo{editor}{E.~Brunskill},
  \bibinfo{editor}{K.~Cho}, \bibinfo{editor}{B.~Engelhardt},
  \bibinfo{editor}{S.~Sabato}, \bibinfo{editor}{J.~Scarlett} (Eds.),
  \bibinfo{booktitle}{International Conference on Machine Learning, {ICML}
  2023, 23-29 July 2023, Honolulu, Hawaii, {USA}}, volume \bibinfo{volume}{202}
  of \textit{\bibinfo{series}{Proceedings of Machine Learning Research}},
  \bibinfo{publisher}{{PMLR}}, \bibinfo{year}{2023}, pp.
  \bibinfo{pages}{22137--22176}.
\bibitem[{Zhang et~al.(2023)Zhang, Zeng, Lin, Xiao, Han, Liu, Sun, and
  Zhou}]{emergent-modularity}
\bibinfo{author}{Z.~Zhang}, \bibinfo{author}{Z.~Zeng},
  \bibinfo{author}{Y.~Lin}, \bibinfo{author}{C.~Xiao},
  \bibinfo{author}{X.~Han}, \bibinfo{author}{Z.~Liu}, \bibinfo{author}{M.~Sun},
  \bibinfo{author}{J.~Zhou}, \bibinfo{title}{Emergent modularity in pre-trained
  transformers}, \bibinfo{year}{2023}. \URLprefix
  \url{https://openreview.net/forum?id=XHuQacT6sa6}.
\bibitem[{Zhang et~al.(2022)Zhang, Lin, Liu, Li, Sun, and Zhou}]{moefication}
\bibinfo{author}{Z.~Zhang}, \bibinfo{author}{Y.~Lin}, \bibinfo{author}{Z.~Liu},
  \bibinfo{author}{P.~Li}, \bibinfo{author}{M.~Sun}, \bibinfo{author}{J.~Zhou},
\newblock \bibinfo{title}{Moefication: Transformer feed-forward layers are
  mixtures of experts},
\newblock in: \bibinfo{editor}{S.~Muresan}, \bibinfo{editor}{P.~Nakov},
  \bibinfo{editor}{A.~Villavicencio} (Eds.), \bibinfo{booktitle}{Findings of
  the Association for Computational Linguistics: {ACL} 2022, Dublin, Ireland,
  May 22-27, 2022}, \bibinfo{publisher}{Association for Computational
  Linguistics}, \bibinfo{year}{2022}, pp. \bibinfo{pages}{877--890}. \URLprefix
  \url{https://doi.org/10.18653/v1/2022.findings-acl.71}.
  \DOIprefix\doi{10.18653/v1/2022.findings-acl.71}.
\bibitem[{Pfeiffer et~al.(2023)Pfeiffer, Ruder, Vulic, and
  Ponti}]{modular-deep-learning}
\bibinfo{author}{J.~Pfeiffer}, \bibinfo{author}{S.~Ruder},
  \bibinfo{author}{I.~Vulic}, \bibinfo{author}{E.~M. Ponti},
\newblock \bibinfo{title}{Modular deep learning},
\newblock \bibinfo{journal}{CoRR} \bibinfo{volume}{abs/2302.11529}
  (\bibinfo{year}{2023}). \URLprefix
  \url{https://doi.org/10.48550/arXiv.2302.11529}.
  \href{http://arxiv.org/abs/2302.11529}{{\tt arXiv:2302.11529}}.
\bibitem[{Pochinkov and Schoots(2023)}]{pochinkov2023dissecting}
\bibinfo{author}{N.~Pochinkov}, \bibinfo{author}{N.~Schoots},
\newblock \bibinfo{title}{Dissecting large language models},
\newblock in: \bibinfo{booktitle}{Socially Responsible Language Modelling
  Research}, \bibinfo{year}{2023}.
\bibitem[{Foster et~al.(2023)Foster, Schoepf, and
  Brintrup}]{selective-synaptic-dampening}
\bibinfo{author}{J.~Foster}, \bibinfo{author}{S.~Schoepf},
  \bibinfo{author}{A.~Brintrup},
\newblock \bibinfo{title}{Fast machine unlearning without retraining through
  selective synaptic dampening},
\newblock \bibinfo{journal}{CoRR} \bibinfo{volume}{abs/2308.07707}
  (\bibinfo{year}{2023}). \href{http://arxiv.org/abs/2308.07707}{{\tt
  arXiv:2308.07707}}.
\bibitem[{Jiang et~al.(2023)Jiang, Sablayrolles, Mensch, Bamford, Chaplot,
  de~las Casas, Bressand, Lengyel, Lample, Saulnier, Lavaud, Lachaux, Stock,
  Scao, Lavril, Wang, Lacroix, and Sayed}]{mistral}
\bibinfo{author}{A.~Q. Jiang}, \bibinfo{author}{A.~Sablayrolles},
  \bibinfo{author}{A.~Mensch}, \bibinfo{author}{C.~Bamford},
  \bibinfo{author}{D.~S. Chaplot}, \bibinfo{author}{D.~de~las Casas},
  \bibinfo{author}{F.~Bressand}, \bibinfo{author}{G.~Lengyel},
  \bibinfo{author}{G.~Lample}, \bibinfo{author}{L.~Saulnier},
  \bibinfo{author}{L.~R. Lavaud}, \bibinfo{author}{M.-A. Lachaux},
  \bibinfo{author}{P.~Stock}, \bibinfo{author}{T.~L. Scao},
  \bibinfo{author}{T.~Lavril}, \bibinfo{author}{T.~Wang},
  \bibinfo{author}{T.~Lacroix}, \bibinfo{author}{W.~E. Sayed},
\newblock \bibinfo{title}{Mistral 7b}  (\bibinfo{year}{2023}).
\bibitem[{Zhang et~al.(2022)Zhang, Roller, Goyal, Artetxe, Chen, Chen, Dewan,
  Diab, Li, Lin, Mihaylov, Ott, Shleifer, Shuster, Simig, Koura, Sridhar, Wang,
  and Zettlemoyer}]{meta-opt}
\bibinfo{author}{S.~Zhang}, \bibinfo{author}{S.~Roller},
  \bibinfo{author}{N.~Goyal}, \bibinfo{author}{M.~Artetxe},
  \bibinfo{author}{M.~Chen}, \bibinfo{author}{S.~Chen},
  \bibinfo{author}{C.~Dewan}, \bibinfo{author}{M.~T. Diab},
  \bibinfo{author}{X.~Li}, \bibinfo{author}{X.~V. Lin},
  \bibinfo{author}{T.~Mihaylov}, \bibinfo{author}{M.~Ott},
  \bibinfo{author}{S.~Shleifer}, \bibinfo{author}{K.~Shuster},
  \bibinfo{author}{D.~Simig}, \bibinfo{author}{P.~S. Koura},
  \bibinfo{author}{A.~Sridhar}, \bibinfo{author}{T.~Wang},
  \bibinfo{author}{L.~Zettlemoyer},
\newblock \bibinfo{title}{{OPT:} open pre-trained transformer language models},
\newblock \bibinfo{journal}{CoRR} \bibinfo{volume}{abs/2205.01068}
  (\bibinfo{year}{2022}). \URLprefix
  \url{https://doi.org/10.48550/arXiv.2205.01068}.
  \href{http://arxiv.org/abs/2205.01068}{{\tt arXiv:2205.01068}}.
\bibitem[{Liu et~al.(2019)Liu, Ott, Goyal, Du, Joshi, Chen, Levy, Lewis,
  Zettlemoyer, and Stoyanov}]{roberta}
\bibinfo{author}{Y.~Liu}, \bibinfo{author}{M.~Ott}, \bibinfo{author}{N.~Goyal},
  \bibinfo{author}{J.~Du}, \bibinfo{author}{M.~Joshi},
  \bibinfo{author}{D.~Chen}, \bibinfo{author}{O.~Levy},
  \bibinfo{author}{M.~Lewis}, \bibinfo{author}{L.~Zettlemoyer},
  \bibinfo{author}{V.~Stoyanov},
\newblock \bibinfo{title}{Roberta: {A} robustly optimized {BERT} pretraining
  approach},
\newblock \bibinfo{journal}{CoRR} \bibinfo{volume}{abs/1907.11692}
  (\bibinfo{year}{2019}). \URLprefix \url{http://arxiv.org/abs/1907.11692}.
  \href{http://arxiv.org/abs/1907.11692}{{\tt arXiv:1907.11692}}.
\bibitem[{Dosovitskiy et~al.(2021)Dosovitskiy, Beyer, Kolesnikov, Weissenborn,
  Zhai, Unterthiner, Dehghani, Minderer, Heigold, Gelly, Uszkoreit, and
  Houlsby}]{vision-transformer-google-2021}
\bibinfo{author}{A.~Dosovitskiy}, \bibinfo{author}{L.~Beyer},
  \bibinfo{author}{A.~Kolesnikov}, \bibinfo{author}{D.~Weissenborn},
  \bibinfo{author}{X.~Zhai}, \bibinfo{author}{T.~Unterthiner},
  \bibinfo{author}{M.~Dehghani}, \bibinfo{author}{M.~Minderer},
  \bibinfo{author}{G.~Heigold}, \bibinfo{author}{S.~Gelly},
  \bibinfo{author}{J.~Uszkoreit}, \bibinfo{author}{N.~Houlsby},
\newblock \bibinfo{title}{An image is worth 16x16 words: Transformers for image
  recognition at scale},
\newblock in: \bibinfo{booktitle}{9th International Conference on Learning
  Representations, {ICLR} 2021, Virtual Event, Austria, May 3-7, 2021},
  \bibinfo{publisher}{OpenReview.net}, \bibinfo{year}{2021}. \URLprefix
  \url{https://openreview.net/forum?id=YicbFdNTTy}.
\bibitem[{Gao et~al.(2020)Gao, Biderman, Black, Golding, Hoppe, Foster, Phang,
  He, Thite, Nabeshima, Presser, and Leahy}]{pile}
\bibinfo{author}{L.~Gao}, \bibinfo{author}{S.~Biderman},
  \bibinfo{author}{S.~Black}, \bibinfo{author}{L.~Golding},
  \bibinfo{author}{T.~Hoppe}, \bibinfo{author}{C.~Foster},
  \bibinfo{author}{J.~Phang}, \bibinfo{author}{H.~He},
  \bibinfo{author}{A.~Thite}, \bibinfo{author}{N.~Nabeshima},
  \bibinfo{author}{S.~Presser}, \bibinfo{author}{C.~Leahy},
\newblock \bibinfo{title}{The {P}ile: An 800gb dataset of diverse text for
  language modeling},
\newblock \bibinfo{journal}{arXiv preprint arXiv:2101.00027}
  (\bibinfo{year}{2020}).
\bibitem[{Russakovsky et~al.(2015)Russakovsky, Deng, Su, Krause, Satheesh, Ma,
  Huang, Karpathy, Khosla, Bernstein, Berg, and Fei-Fei}]{imagenet}
\bibinfo{author}{O.~Russakovsky}, \bibinfo{author}{J.~Deng},
  \bibinfo{author}{H.~Su}, \bibinfo{author}{J.~Krause},
  \bibinfo{author}{S.~Satheesh}, \bibinfo{author}{S.~Ma},
  \bibinfo{author}{Z.~Huang}, \bibinfo{author}{A.~Karpathy},
  \bibinfo{author}{A.~Khosla}, \bibinfo{author}{M.~Bernstein},
  \bibinfo{author}{A.~C. Berg}, \bibinfo{author}{L.~Fei-Fei},
\newblock \bibinfo{title}{{ImageNet Large Scale Visual Recognition Challenge}},
\newblock \bibinfo{journal}{International Journal of Computer Vision (IJCV)}
  \bibinfo{volume}{115} (\bibinfo{year}{2015}) \bibinfo{pages}{211--252}.
  \DOIprefix\doi{10.1007/s11263-015-0816-y}.
\bibitem[{He et~al.(2016)He, Zhang, Ren, and Sun}]{resnet-original-paper}
\bibinfo{author}{K.~He}, \bibinfo{author}{X.~Zhang}, \bibinfo{author}{S.~Ren},
  \bibinfo{author}{J.~Sun},
\newblock \bibinfo{title}{Deep residual learning for image recognition},
\newblock in: \bibinfo{booktitle}{IEEE conference on computer vision and
  pattern recognition}, \bibinfo{year}{2016}, pp. \bibinfo{pages}{770--778}.
\bibitem[{Thorpe and van Gennip(2018)}]{residual-deep-limits}
\bibinfo{author}{M.~Thorpe}, \bibinfo{author}{Y.~van Gennip},
\newblock \bibinfo{title}{Deep limits of residual neural networks},
\newblock \bibinfo{journal}{arXiv preprint arXiv:1810.11741}
  (\bibinfo{year}{2018}).
\bibitem[{Heimersheim and Turner(2023)}]{residual-grows}
\bibinfo{author}{S.~Heimersheim}, \bibinfo{author}{A.~Turner},
  \bibinfo{title}{Residual stream norms grow exponentially over the forward
  pass},
  \bibinfo{howpublished}{\url{https://www.alignmentforum.org/posts/8mizBCm3dyc432nK8/residual-stream-norms-grow-exponentially-over-the-forward}},
  \bibinfo{year}{2023}. \bibinfo{note}{14 min read}.
\bibitem[{Wang et~al.(2022)Wang, Variengien, Conmy, Shlegeris, and
  Steinhardt}]{circuits-ioi-interpretability-redundancy}
\bibinfo{author}{K.~Wang}, \bibinfo{author}{A.~Variengien},
  \bibinfo{author}{A.~Conmy}, \bibinfo{author}{B.~Shlegeris},
  \bibinfo{author}{J.~Steinhardt},
\newblock \bibinfo{title}{Interpretability in the wild: a circuit for indirect
  object identification in gpt-2 small},
\newblock \bibinfo{journal}{arXiv preprint arXiv:2211.00593}
  (\bibinfo{year}{2022}).
\bibitem[{McGrath et~al.(2023)McGrath, Rahtz, Kram{\'{a}}r, Mikulik, and
  Legg}]{hydra-effect}
\bibinfo{author}{T.~McGrath}, \bibinfo{author}{M.~Rahtz},
  \bibinfo{author}{J.~Kram{\'{a}}r}, \bibinfo{author}{V.~Mikulik},
  \bibinfo{author}{S.~Legg},
\newblock \bibinfo{title}{The hydra effect: Emergent self-repair in language
  model computations},
\newblock \bibinfo{journal}{CoRR} \bibinfo{volume}{abs/2307.15771}
  (\bibinfo{year}{2023}). \URLprefix
  \url{https://doi.org/10.48550/arXiv.2307.15771}.
  \DOIprefix\doi{10.48550/arXiv.2307.15771}.
  \href{http://arxiv.org/abs/2307.15771}{{\tt arXiv:2307.15771}}.

\end{thebibliography}

Appendix

\section{Appendix}

\subsection{Resampling Methods}\label{app:resampling-methods}

As we are not performing resampling based on a single possible input, but rather comparing many methods, we seek a few possible inputs to gain an in-distribution sample of what model activations might look like. We consider three methods. The first is to generate a string of random alphanumeric characters. The second is to generate a random selection of possible input tokens from the model. The last is to attempt to generate a random in-distribution text based on a single token. We see some extracts in Table \ref{tab:random-text-examples}. The three methods are plotted together, as their performances were relatively similar.

\begin{table}[h]
    \caption{Extracts of the inputs for text-based resampling.}
    \centering
    \vspace{4px}
    \begin{tabular}{cc}
        \toprule
         Resample method & Text extracts \\
         \midrule
         Random characters &\ \texttt{3KhCfVVkC5VNSchVtAyKurMGCHaMHS25SSKrJRD6yaNZSwSN3}...
         \\
         Random token IDs & \ \texttt{ built Pilot Citiesdamage UK honouredoped apples}...\\
         Random generated & \ \texttt{  That's what I'm going to do.\textbackslash nIt was pretty bad}... \\
         \bottomrule
    \end{tabular}
    \label{tab:random-text-examples}
\end{table}

\end{document}